\documentclass[lettersize,journal]{IEEEtran}
\usepackage{amsmath,amsfonts}
\usepackage{algorithmic}
\usepackage{algorithm}
\usepackage{array}
\usepackage{textcomp}
\usepackage{stfloats}
\usepackage{url}
\usepackage{verbatim}
\usepackage{graphicx}
\usepackage[backref]{hyperref}
\usepackage{graphicx}
\usepackage{xcolor}
\usepackage[font=footnotesize]{caption}
\usepackage{tikz}

\newcommand\copyrighttext{%
  \footnotesize \textcopyright 2023 IEEE. Personal use of this material is permitted. Permission from IEEE must be obtained for all other uses, in any current or future media, including reprinting/republishing this material for advertising or promotional purposes, creating new collective works, for resale or redistribution to servers or lists, or reuse of any copyrighted component of this work in other works.
}
\newcommand\copyrightnotice{%
\begin{tikzpicture}[remember picture,overlay]
\node[anchor=south,yshift=10pt] at (current page.south) {\fbox{\parbox{\dimexpr\textwidth-\fboxsep-\fboxrule\relax}{\copyrighttext}}};
\end{tikzpicture}%
    }

\usepackage[T1]{fontenc} 
\usepackage{booktabs} 

\begin{document}

\title{Interpretable Learned Emergent Communication\\for Human-Agent Teams}


\author{Seth Karten,
Mycal Tucker,
Huao Li,
Siva Kailas,
Michael Lewis,
and Katia Sycara
\thanks{Seth Karten, Siva Kailas, and Katia Sycara are
with the Robotics Institute, Carnegie Mellon University, Pittsburgh, PA
15260 USA (e-mail: skarten@cs.cmu.edu; skailas@cs.cmu.edu; katia@cs.cmu.edu).}%
\thanks{Mycal Tucker is with the Department of Aeronautics and Astronautics, Massachusetts Institute of Technology, Cambridge, MA 02139 USA (e-mail: mycal@csail.mit.edu).}%
\thanks{Huao Li and Michael Lewis are with the School of Information
Science, University of Pittsburgh, Pittsburgh, PA 15260 USA (e-mail:
hul52@pitt.edu; ml@sis.pitt.edu).}
}
\markboth{Preprint}
{Karten \MakeLowercase{\textit{et al.}}: Interpretable Communication for Human-Agent Teams} 


\maketitle

\copyrightnotice

\begin{abstract}
Learning interpretable communication is essential for multi-agent and human-agent teams (HATs). In multi-agent reinforcement learning for partially-observable environments, agents may convey information to others via learned communication, allowing the team to complete its task.
Inspired by human languages, recent works study discrete (using only a finite set of tokens) and sparse (communicating only at some time-steps) communication. However, the utility of such communication in human-agent team experiments has not yet been investigated. In this work, we analyze the efficacy of sparse-discrete methods for producing emergent communication that enables high agent-only and human-agent team performance. We develop agent-only teams that communicate sparsely via our scheme of  Enforcers that  sufficiently constrain communication to any budget. Our results show no loss or minimal loss of performance in benchmark environments and tasks. In human-agent teams tested in benchmark environments,  where agents have been modeled using the Enforcers,  we  find that a prototype-based method produces meaningful discrete tokens that enable human partners to learn agent communication faster and better than a one-hot baseline. Additional HAT experiments show that an appropriate sparsity level lowers the cognitive load of humans when communicating with teams of agents and leads to superior team performance.

\end{abstract}


\section{Introduction}

\IEEEPARstart{M}{ulti-agent} reinforcement learning (MARL) has been successfully applied in a variety of multiplayer games~\cite{gronauer2021multi,zhang2021multi}, but trained agents often only collaborate well with humans in specific settings. For example, in StarCraft~\cite{starcraft} and Dota 2~\cite{dota2}, teams of agents may be trained to compete against humans; or in Hanabi~\cite{siu2021evaluation} and Overcooked~\cite{carroll2019utility}, only a single agent may cooperate with a human. Prior art in training agents to collaborate in more complex settings that require communication among teammates (e.g., blindly crossing a traffic junction) has considered agent-only teams~\cite{commnet,ic3net}, but translating such techniques to support human teammates, especially without human data during training, remains challenging~\cite{andreas2017translating}.

\begin{figure}[!t]
    \centering
    \includegraphics[width=\columnwidth]{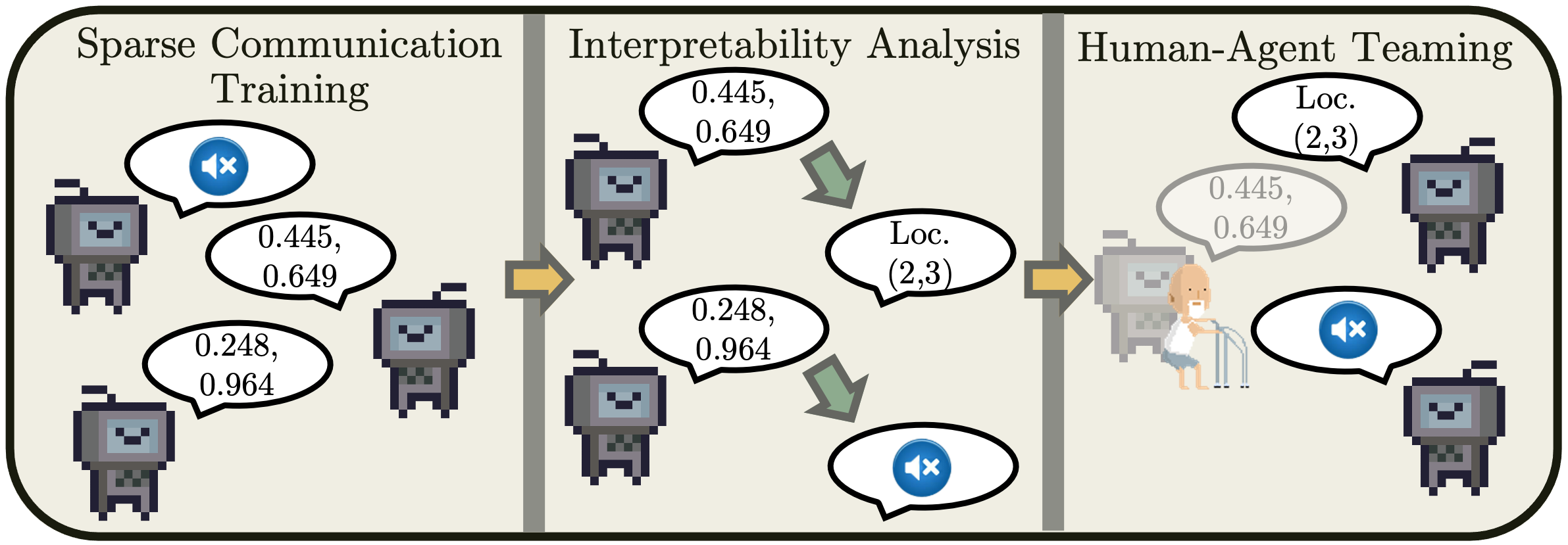}
    \caption{Our method has three phases: During sparse communication training, we train agents through self-play to learn an emergent message paradigm and to communicate infrequently according to a budget. In the interpretability analysis, messages are analyzed to determine the observations that they encode. Additional messages are removed in case they contain no information. Finally, in human-agent teaming, a ghost agent decodes the information and sends encoded messages for the human, who chooses the action.}
    \label{fig:fig1}

\end{figure}

Teaming with artificial agents is generally novel for humans; therefore, teaming requires a learning process for humans to adapt to new protocols for communication and collaboration. 
The dissimilarity of decision-making and lack of co-training~\cite{nikolaidis2013} introduces a significant challenge for learning a shared `language' for both humans and agents. A large body of research investigates the problems of translating natural language into a form usable by agents and generating intelligible replies in return.  Our work takes the complementary approach of searching for ways to make it easier for humans to communicate with agents in their own language.
Biases in human cognition are a product of both neural substrates~\cite{soltani2016} and evolutionary processes~\cite{kliegr2021}, which predispose humans to perceive and think in particular ways and not in others.  Policies learned in the absence of human data, such as unsupervised interaction among agents, are less likely to conform to human  biases than approaches that incorporate human inputs.  Developers of AlphaGo, a family of Go playing programs that outperform human experts, found that while versions trained through supervised learning were superior in predicting the moves of human experts, versions trained exclusively through self-play were much better players~\cite{silver2017}.  These self-trained programs had perfect records against human opponents, found new joseki (corner sequences) unknown to human players~\cite{silver2017}, and were described by human players as “amazing, strange or alien”~\cite{chan2017}.  A sister StarCraft playing program, AlphaStar~\cite{vinyals2019},  followed the opposite tack by learning from replays of human matches and limiting speed and observations to human-like values~\cite{vinyals2019}.  In this case, despite reaching Grandmaster status (99.8\%), the AlphaStar agents team could be defeated and was described as “like it is playing a ‘real’ game of StarCraft and doesn’t completely throw the balance off by having unrealistic capabilities”~\cite{vinyals2019}.  Human difficulty in understanding behavior developed through self-play can be seen in even very simple games.  Despite providing saliency maps and/or reward decomposition graphics for offering extra insight into the factors contributing to an action, observers could not predict the next action in Ms. Pac-Man~\cite{iyer2018transparency} or a drastically simplified real-time strategy game~\cite{anderson2019} at better than chance levels.

During human-agent teaming, humans often struggle to understand the intent of agent partners, which is necessary for an effective partnership.  Recent work has shown that bidirectional human-agent communication is sometimes required for peak performance in human-agent teaming~\cite{marathe2018bidirectional}.
While prior work in emergent communication establishes how agents may learn to communicate to accomplish their goals, the learned communication conventions often exhibit undesirable properties and fail to conform to human cognitive mechanisms.
For example, emergent communication is often continuous, while humans communicate in natural language, which consists of discrete linguistic tokens or prototypes. 
Furthermore, humans learn from few examples and use compositional language that encompasses more complex meanings than was demonstrated to them~\cite{lake2019human}. 
Prior art has taken initial steps towards closing the gap between emergent and human communication: in agent-only teams in specific scenarios, discrete communication can perform as well as continuous communication~\cite{li2021learning}, and in our prior work, we showed how learning discrete prototypes promotes robustness in noisy communication channels, as well as human interpretability  and zero-shot generalization~\cite{discreteComm}. 
However, learning suitable discrete communication protocols that exhibit high performance in achieving team goals in sequential team decision-making remains a big challenge.

A recent meta-analysis~\cite{marlow2018does} found that communication quality (rating as “effective and clear”) had a significantly more substantial relationship with performance than the frequency of communication in human teams. 
In fact, excessive communication can present problems both in recognizing critical information and in remembering it. Recognizing an actionable message from a background of irrelevant ones is a classic signal detection problem~\cite{GreenSwets66} for which misses are known to rise as the number of irrelevant messages increases. Cognitive load theory~\cite{van2005cognitive} suggests that a large volume of communication may also interfere with the process of transferring information from working memory to long-term memory, which creates difficulty in remembering previously received information accurately. Thus, high frequency/high redundancy communication is likely to harm humans' abilities to recognize and learn from `high quality' communication.

As an attempt to make agent communication akin to human communication in agent self-play, sparsity constraints have been shown to reduce the total amount of communication. Sparsity aims to minimize the number of unnecessary communications between agents to adhere to bandwidth constraints. Previous methods attempt to minimize the total communication while minimizing the loss of performance~\cite{wang2020learning}. However, achieving this is very challenging: works that use gating (a function that determines whether to pass a message or not) often exhibit high variance and tend not to converge to the optimal communication budget~\cite{agrawal2021learning}. Our Enforcers scheme builds upon gating methods but reduces variance sufficiently to converge to the optimal gating value by using a soft threshold.

To the best of our knowledge, this is the first work investigating the intelligibility of agent-generated communication models and their effects on performance in human-agent teams. 
We train our agents to learn human interpretable, discrete prototypes to interpret a message's intent by the receiver. The agent self-play method also constrains communication and considers the effects of different communication budgets on communication robustness and team performance. We develop an emergent communication interpretability scheme to transition from an agent-only communication space to a human-interpretable interface. In human-agent communication experiments, we explore  the efficacy of humans learning to communicate with agents using discrete prototypes compared to a discrete 1-hot representation.\footnote{A one-hot is a group of bits among which the legal combinations of values are only those with a single high (1) bit and all the others low (0).    In statistics, dummy variables represent a similar technique for representing categorical data.}

Human-agent teaming experiments test two hypotheses. The first is whether humans can learn to communicate in Human-Agent Teams (HATs) using the emergent discrete prototypes from agent self-play. This is tested with single-agent HATs in the Parent and Lost Child environment, a variant of Predator-Prey. The second hypothesis is that an appropriate level of sparsity in communication reduces the cognitive load of human teammates and leads to better team performance. This is tested with multi-agent HATs in a blind Traffic Junction environment.

Our  work additionally analyzes the effects of sparsity at various communication  budgets to find the optimal minimum budget for human-agent teams through cognitive load and  task performance.

Our contributions are as follows:

\begin{itemize}

\item We propose a novel MARL method that uses an interpretability analysis to produce sparse-discrete communication in HAT. We evaluate its efficacy in human subject experiments in HATs with (a) single human-single agent in different roles and (b) single human-multiple agents. 

\item Our results indicate that humans can learn the emergent discrete prototype `language' generated by agent self-play faster as compared to a baseline. 

\item This work is the first  that focuses not only on learning human interpretable communication but, most crucially, on the effects of learned sparse communication in MARL on HAT \textit{performance and human  workload}. In particular, we expand previous findings about communication sparsity from human-human teams to human-agent teams. Our results show that an appropriate frequency level leads to the best team performance and the lowest cognitive load of human teammates.

\end{itemize}

\section{Related Work}\label{related_work}
Our work studies multi-agent reinforcement learning at the intersection of research in sparse and discrete emergent communication, evaluated in the context of human-agent teams.

\subsection{Multi-Agent Reinforcement Learning}
Multi-agent reinforcement learning studies a team of agents working to maximize shared performance on a task. By learning to communicate through backpropagation, agents can learn task-specific coordination directly from error derivatives. Typically, a centralized training, decentralized execution paradigm enables agents to learn from privileged information but act independently~\cite{foerster2016learning}. 
Instead, we use a fully decentralized training setup with shared parameters to foster faster training in cooperative tasks~\cite{commnet}.

\subsection{Sparse Communication}
In the context of multi-agent systems, sparse communication necessitates limiting the total communication exhibited between agents, measured by the number of bits sent.
Sparse communication has been explored by learning a communication gate, learning whom to target, and compressing communication tokens.
Gating methods use a neural network layer to learn a gating function to decide whether to pass through the message~\cite{mao2020learning,vijay2021minimizing}. 
Targeting methods learn who to send messages and/or who would be receiving messages~\cite{tarmac,graphMA,goyal2020variational,kim2019learning}.
Information bottleneck methods attempt to minimize the entropy of messages between agents to learn whom to target communications with a centralized communication targeting system~\cite{rashid2018qmix,wang2020learning}.  These methods try only to remove unnecessary communication, but they have been shown to decrease the overall reward, which leads to suboptimal task performance.
 Targeting methods work as a complementary model to gating and information bottleneck methods. Thus, this may be used as an additional model to reduce communication further. All these methods only have one suboptimal budget to reduce communication. 
  Compression is enabled by limiting the size of communication tokens to a fixed size or a bitwise encoding~\cite{seraj2022learning,freed2020sparse} though recent work has shown that continuous encodings can contain more information~\cite{discreteComm}.
Our work builds upon gating methods but reduces variance sufficiently to converge to the optimal gating value by using a soft threshold.
We are able to converge to various communication budgets, allowing our models to learn communication conventions that complement human preferences while maintaining multi-agent system performance. Our paper also analyzes the performance of sparsity in human trials.

\subsection{Interpretable Communication Formats}
Regardless of communication frequency, researchers have developed a variety of communication formats in an effort to improve human interpretability.
For example, inspired by the discrete nature of words in human language, one work discretizes emergent communication by forcing agents to communicate via one-hot or binary vectors~\cite{lowe2017multi}.
Other works focus on the compositionality of tokens to create simplified ``sentences''~\cite{mordatch2018emergence} or train agents to communicate directly via natural language~\cite{peterson2017adapting,visdial,agarwalvisdial2019}.
Unfortunately, agents trained by such techniques often perform worse in human trials than in multi-agent teams, indicating that the communication interpretability remains limited~\cite{lazaridou2016multi}.
Other work decides the interpretation of messages based on the effect on a human listener~\cite{andreas2017translating}.
In our work, we train agents to communicate via a discrete set of tokens in a continuous space~\cite{discreteComm}. 
While maximally informative to the agents, these sets of tokens may conflate intent with location or relations among items in correlated but indirect ways. Our experiments test the degree to which discretization and sparsity can overcome potential cognitive incompatibilities and allow humans to exploit the semantic richness of agent token sets.  In our single-trial experiments, participants must learn the semantics of tokens while using them to accomplish their joint tasks.  The one-hot encoding provides an isomorphic mapping, known to be compatible with human cognition, but must be learned token by token as a paired-associate task. By contrast, embedding spaces learned by the agents provide a relational structure among tokens, reducing the learning that must occur~\cite{tse2007schemas}. 
\begin{figure}[!t]
    \centering
    \includegraphics[width=.8\columnwidth]{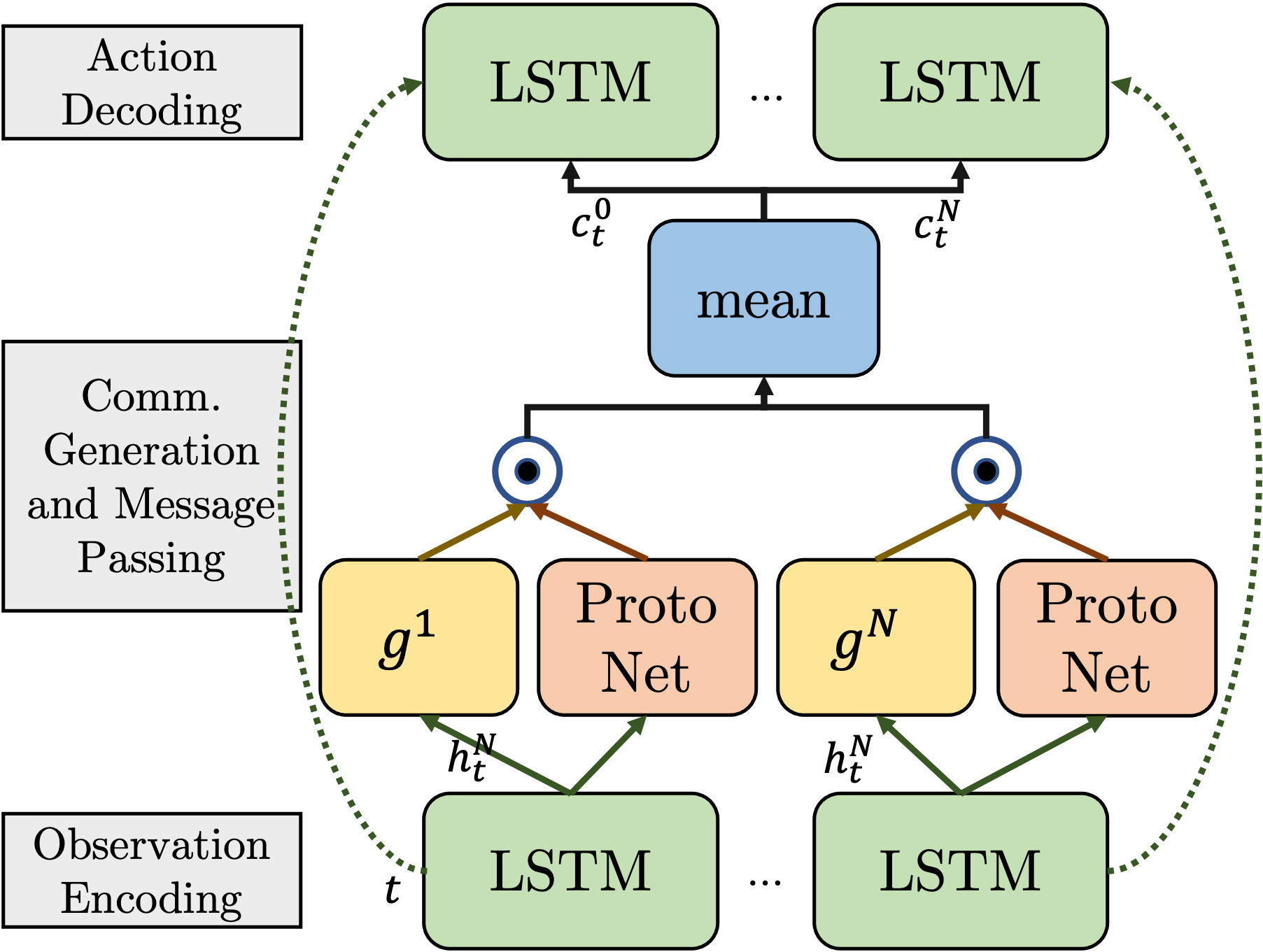}
    \caption{Above is the Enforcers architecture, including the MARL with communication pipeline.}
    \label{fig:protoNet}
\end{figure}
\subsection{Human-Agent Teaming}
Most human-agent teaming has focused on adapting policies to allow humans and agents to understand their partners' intent, which is necessary for effective partnership.
Both human and agent adaptation benefits team performance~\cite{li2021individualized}.
Some agents are able to recognize changes in human intent given observations of their actions. The agents can then adapt their policy to coordinate better with the human~\cite{hughes2020inferring}.
Agents have also been trained to learn the effects of their actions on other agents. When combined with communication, this has led to increased coordination of agent-only teams~\cite{jaques2019social}.
These works assume agents will adapt to their teammates' policies by observing their actions.
Instead, given the importance of bidirectional human-agent communication in team performance, we focus on enabling humans and agents to also adapt based on \textit{communication}~\cite{marathe2018bidirectional}.

Recent works have begun to explore communication in human-agent teams. 
In agent self-play and human-agent teaming, sharing the intent or goal was most useful towards increasing performance~\cite{li2015communication}. However, this study only involved tasks with simplistic agents who follow hard-coded randomized paths, in which communication was not required to successfully complete the task.
Other work has built upon this by introducing tasks that may require communication. They found that communicating both beliefs and goals, while minimizing communication frequency, improved human-agent teaming~\cite{van2020learning}, but worked with simplistic agents that are given human-designed observation information to complete their task. They then learn to recognize when they need additional information and learn to communicate from a list of predefined goals.
In our work, we use state-of-the-art agents to team with humans using \textit{emergent communication}. Our agents learn both communication and action policies. In our benchmarks, communication is necessary to behave optimally, including cases when a single human must coordinate with multiple agents.

\section{Agent Self-Play}\label{agent_training}

\begin{table*}[t!]
\centering
\begin{tabular}{c|c c c c | c c c}
\specialrule{.2em}{.1em}{.1em}
& \multicolumn{4}{c}{\textbf{Traffic Junction}} \vline & \multicolumn{2}{c}{\textbf{Predator-Prey Cooperative}}\\
 \specialrule{.2em}{.1em}{.1em}
 & \multicolumn{2}{c}{\textbf{Easy}} \vline & \multicolumn{2}{c}{\textbf{Medium}} \vline & \textbf{5$\times$5, N=3} & \textbf{10$\times$10, N=5}\\
  \textbf{Model} & Convergence Epoch  & Success \% &  Convergence Epoch & Success \% & \multicolumn{2}{c}{Average Rewards}\\
 \hline
 Fixed-Cts & 101 & .993 & 675 & .997 & 2.25 & 7.64 \\ 
 Fixed-Proto & 199 & .993 & 927 & .959 & 2.18 & 7.13 \\ 
 Gated-Cts & 652 & .968 & 1320 & .926 & 1.38 & 6.51\\ 
 Gated-Proto & 1262  & .977  & 1518 &  .920 & 0.94 & 5.69\\
 \textbf{Enforcer-$b^*$} & +35 & .983 & +196 & .947 & 2.07 & 7.22\\
\specialrule{.1em}{.05em}{.05em}
\end{tabular}
\caption{\textbf{Traffic Junction:} In two Traffic Junction environments, we compared the convergence epoch and success rate for fixed (at all time-steps) vs. gated (sparse) communication and continuous vector vs. prototype-based (discrete) tokens. Prototype-based agents achieve a similar success rate to continuous communication agents. The Enforcer method can stably maximize success while using optimal communication $b^*$ and discrete prototypes with a few additional epochs from Fixed-Proto. \textbf{Predator-Prey:} In two cooperative predator-prey environments, we measured the team reward for agents using fixed vs. gated communication and continuous vector vs. prototype-based (discrete) tokens. Unlike naive gating approaches, the Enforcer method stably maximizes reward while using optimal communication $b^*$ and discrete prototypes.}
\label{table:TJ_Full}
\end{table*}

In this section, we introduce the Enforcers, a curriculum of constraints necessary for enabling stable multi-agent sparse-discrete emergent communication. 
Agents are trained exclusively with other agents. 
In agent-only experiments, we show that our models can perform competitively with respect to other agent-only models on benchmark multi-agent tasks and environments that utilize communication. 
Additionally, our method can train agents to adhere to various communication budgets from maximum communication (100\% of the time) to the optimal communication budget for a task (e.g., only communicating 10\% of the time), which is assessed in benchmark environments: Predator-Prey and Traffic Junction.

\begin{table}[t!]
\centering
\begin{tabular}{c|c c}
\specialrule{.2em}{.1em}{.1em}
\multicolumn{3}{c}{\textbf{Soft Enforcer (Comm \% $\backslash$ Success \%)}}\\
 \specialrule{.2em}{.1em}{.1em}
  \textbf{Budget} &  \textbf{Easy} &  \textbf{Medium} \\
 \hline
 100 & - $\backslash$ .993 & - $\backslash$ .959 \\
 90 & .889 $\backslash$ .983 & .883 $\backslash$ .955 \\ 
 70 &  .781 $\backslash$ .986 &  .682 $\backslash$ .946 \\
 50 & .588 $\backslash$ .980  & .445 $\backslash$ .947 \\ 
 30 & .282 $\backslash$ .983  & .261 $\backslash$ .931 \\
 20 & .195 $\backslash$ .959 & - $\backslash$ - \\
\specialrule{.1em}{.05em}{.05em}
\end{tabular}
\caption{\textbf{Traffic Junction:} The table above shows the training results using the  Enforcers with various budgets $b$. The fraction of total communication $\backslash$ success rate is compared for each imposed budget. The method is able to yield consistent performance while an optimal budget is observed.}
\label{table:tj_soft}
\end{table}

\subsection{Problem Setup}
We formulate our multi-agent problem as a decentralized, partially observable Markov Decision Process with communication (Dec-POMDP-Comm). Each agent or human receives a partial observation of the environment, so as a team, they must learn to communicate essential information to complete the task adequately. Additionally, we require \textit{sparse} communication: each agent must minimize total communications according to a communication budget $b$. First, define $B$ as the total number of bits communicated if an agent emits a communication vector at each time-step. We define $b = \frac{B}{t_\delta}$ as the average number of bits communicated over any contiguous subset of the episode $t_\delta$. For ease of analysis, we define $B = |\mathcal{T}| * |c|$ as the length of the episode times the size in bits of the communication vector, which makes $b \in [0,1]$. We encode this into the reward function in section~\ref{methodology}.
Our problem is formulated by the tri-objective of discovering the optimal constrained communication-action policy. That is, the agents must learn to 1) communicate effectively, 2) act effectively, and 3) obey communication sparsity constraints. Communications occur at discrete, uniform time-steps.

Formally, our problem is defined by the 8-tuple, $(\mathcal{S},\mathcal{A},\mathcal{C},\mathcal{T},\mathcal{R},\mathcal{O},\Omega,\gamma)$. We define $\mathcal{S}$ as the set of states, $\mathcal{A}_i \, , \, i\in[1,N]$ as the set of actions, which includes task-specific actions, and $\mathcal{C}_i$ as the set of communications for $N$ agents.  $\mathcal{T}$ is the transition between states due to the multi-agent joint action space $\mathcal{T}: \mathcal{S} \times \mathcal{A}_1,...,\mathcal{A}_N \to \mathcal{S}$. $\Omega$ defines the set of observations in our partially observable setting. Partial observability requires communication to complete the tasks successfully. $\mathcal{O}_i: \mathcal{C}_1,...,\mathcal{C}_N \times \mathcal{S} \to \Omega$ maps the communications and state to a distribution of observations for each agent. $\mathcal{R}$ defines the reward function and $\gamma$ defines the discount factor.
We build on the objective in \cite{discreteComm}, in which we aim to maximize the total expected reward of all agents, as follows,
\begin{equation*}
    \max\limits_{\pi: \mathcal{S} \to \mathcal{A} \times \mathcal{C}} \mathbb{E}  \sum_{t \in \mathcal{T}} \sum_{i \in N} \gamma \mathcal{R}(s_t, a_t) \end{equation*}
such that, $(a_t, c_t) \sim \pi$, $s_t \sim \mathcal{T}(s_{t-1})$.

We use REINFORCE \cite{williams1992simple} to train both the gating function and policy network subject to the previous constraints. In order to calculate the information similarity, we compute loss by comparing each agent's decoded state against the entire state but enforce decentralized execution and testing.

\subsection{Methodology}\label{methodology}

The MARL with communication pipeline consists of an observation encoding step, a message passing step, and an action decoding step. 
Similar to IC3Net~\cite{ic3net}, each agent uses a recurrent encoding and decoding step. However, our model novelly addresses the communication generation and message passing step. As depicted in Fig.~\ref{fig:protoNet}, during
each time-step, each agent receives an observation, which is passed into each agent's LSTM\footnote{Long Short-term Memory is a Recurrent Neural Network (RNN) architecture  }. The hidden state is passed to the gating function and the prototype network. Rather than a probabilistic gate in IC3Net, the Enforcers' gating function decides whether to pass a message to other agents based on the latent state information. The prototype network receives continuous hidden state information. The prototype network encodes and compresses the relevant observation and intent/coordination information into one of a discrete set of emergent prototype vectors. We then take the Hadamard product between the gating value and discrete prototype vectors, which masks communication output according to the gating value. These communication messages are passed to other agents, where each agent takes the mean of the messages received. This value is passed to an agent's LSTM, which, finally, produces an action.

The construction of the communication message is similar to an unsupervised learning objective, which determines useful latent information to pass to other agents based on the policy loss gradient. 
Due to the high variance of dual communication and action policy learning, a communication curriculum is applied. A success criterion must be achieved at each level of the curriculum to advance to the next phase. Let $\lambda$ define some hyperparameter. The communication curriculum changes the reward, 
\begin{equation*}
    R^i = R^i_{env} - \lambda R^i_{comm},
\end{equation*}
at each phase, where $ R^i_{comm}$ is defined during each phase. During the “Open Gate” phase, communication is hard-coded to always occur and $ R^i_{comm} = 0$. After achieving the success criterion, the “Positive Communication Reward” phase rewards communication with $ R^i_{comm} = |1-c|$. This phase reduces instability when transitioning from an open gate to a learned gating function. After achieving the same success criterion with the new constraint, the Enforcers phase is implemented.

The Enforcers apply a soft constraint to the reward function to ensure that the communication is within a budget. The communication reward penalty's main purpose is to constrain communication for the total budget and prevent communication bursts. This helps the reward-based sparse method achieve a particular sparse budget.
Define a reward penalty proportional to a scaled distance to the proposed fraction of the total budget $b$ for some observed fraction of total communication $c$,
\begin{equation*}
    R^i_P = \begin{cases} 
      \frac{b-c}{b} & c \leq b\\
      \frac{b-c}{1-b} &  c > b
   \end{cases},
\end{equation*}
We also incorporate a first-order derivative term, 
\begin{equation*}
    R^i_D = R^i_P - R^{i-1}_P,
\end{equation*}
and an integral term, 
\begin{equation*}
    R^i_I = \sum_{j=0}^i R^j_P,
\end{equation*}
to the communication penalty, 
\begin{equation*}
    R^i_{commSoft} = \lambda_P R^i_P + \lambda_D R^i_D + \lambda_I R^i_I.
\end{equation*}
The hyperparameters are tuned empirically. Note that the method limits the integral term such that $|R^i_I | < K$ for some hyperparameter $K$ for stability.
Even with our optimizations, we find reward-based sparsity methods to be high variance. We find that the hyperparameter ranges are binary in that only carefully determined ranges work. Outside these ranges, there is no convergence to the sparse budgets. We use $\lambda_P = 1., \lambda_D = 1.6, \lambda_I = 0.026, K=50$.

\begin{figure}
    \centering
    \includegraphics[width=.4\columnwidth]{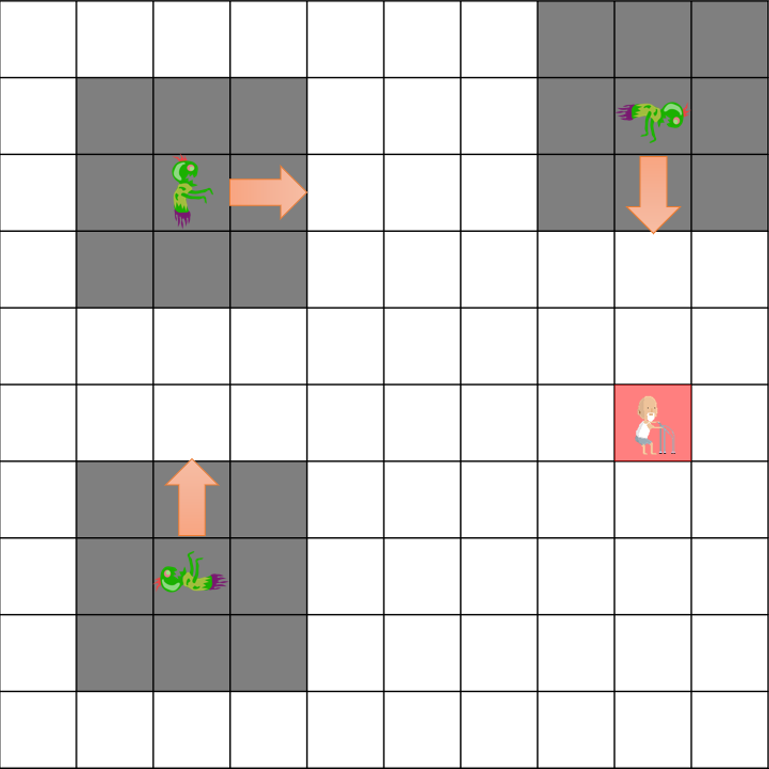}
    \includegraphics[width=.4\columnwidth]{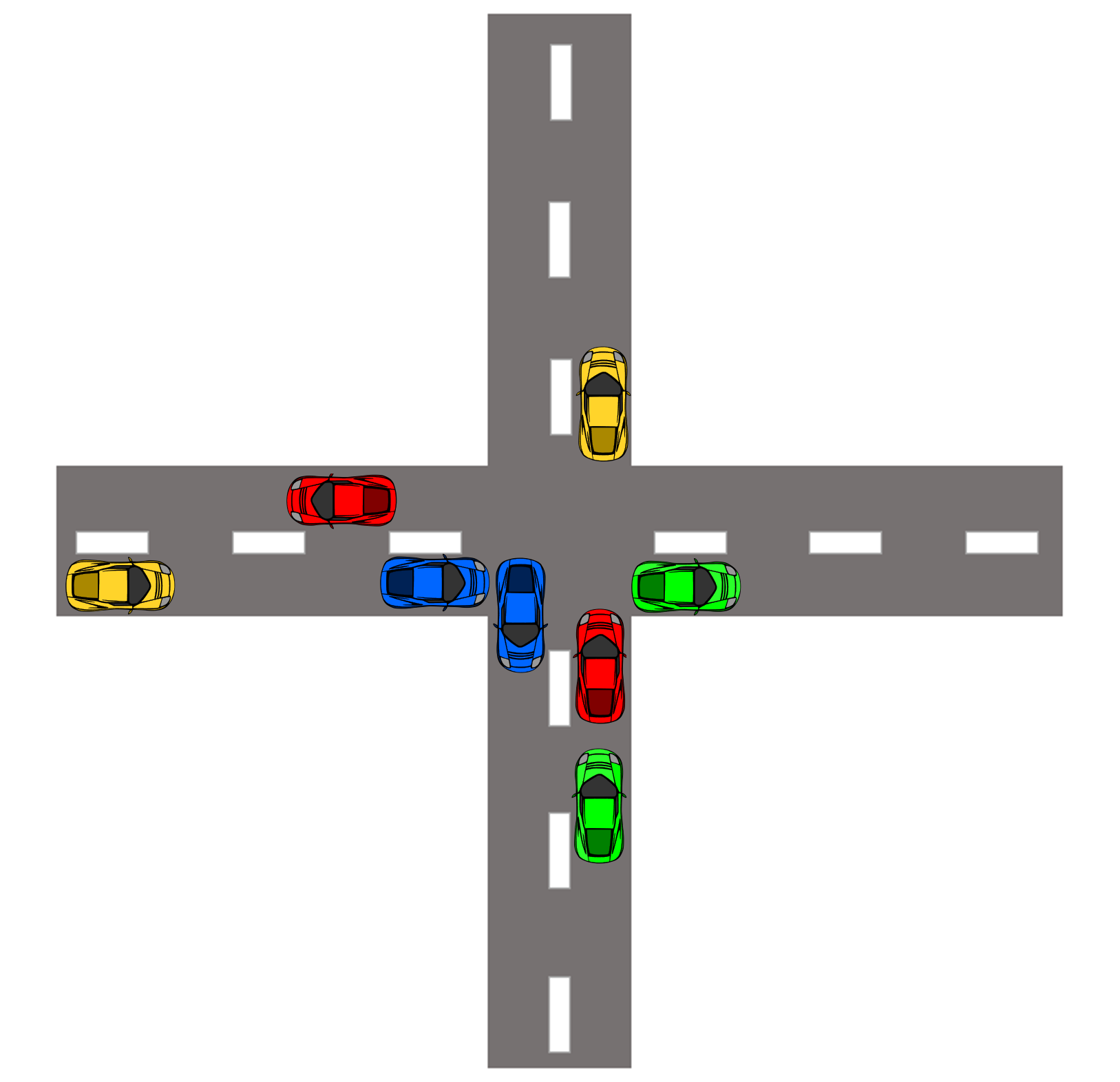}
    \caption{\textbf{Left:} $10\times 10$ Predator-Prey. The human in the red square denotes the prey, while the remaining entities denote the predators. The arrows denote the actions taken by each predator, and the gray shading denotes the vision of the predators.
    \textbf{Right:} Two-lane bidirectional traffic junction.
    The agents enter the junction, driving on the right lane from any of the four entrances randomly. The agent's path entails either going straight at the intersection, turning left at the intersection, or turning right at the intersection. The agents then directly proceed to exit the intersection.
    }
    \label{fig:tj_env}

\end{figure}

\subsection{Benchmark Agent-Only Experiments}
\subsubsection{Experimental Setup}\label{sec:experimental_setup}
We tested the Enforcers scheme in Predator-Prey and Traffic Junction benchmark environments~\cite{ic3net}.
In Predator-Prey, a precursor to the Human and Lost Child scenario, $N$ predator-agents search for one prey agent and then travel to its location. This is a fully-cooperative scenario, so ideally, the prey learns to communicate its location to allow for optimal navigation of the predators. Predator and prey agents each have a uniformly random chance of spawning in any cell in the grid at the beginning of the episode before searching for $T$ time-steps ($5\times 5$: $T=20$; $10\times 10$: $T=40$).

In Traffic Junction, up to 10 agents navigate a two-lane bidirectional traffic junction as shown in Fig.~\ref{fig:tj_env}. The agents are unable to observe each other; they are \textit{"blind"}, so they must communicate to avoid collisions both in the junction and within a given lane (e.g., if the front agent brakes). Agents are spawned in the environment with probability $p$ at each time-step for a fixed total number of time-steps $T$ (easy: $p=0.1$, $T=20$; medium: $p=0.05$, $T=40$).

\subsubsection{Results}
Setting a budget with the Enforcers, the model can converge to various communication budgets, including the optimal communication budget $b^*$. Performance is evaluated over three seeds. The Enforcer method is able to yield performance equivalent to unconstrained methods, i.e., continuous communication,  while exhibiting discrete-sparse properties. Table~\ref{table:TJ_Full} shows that our method, \textbf{Enforcer}-$b^*$, is able to converge to a high success rate with few additional training epochs after introducing the Enforcer constraints on the Fixed-Proto method. Table~\ref{table:tj_soft} shows that the method can constrain communication until an optimal budget threshold with nearly no change in reward. Overall, the Enforcer method is able to provide high task performance at various communication budgets with limited additional training.

\begin{figure*}[!t]
    \centering
    \includegraphics[width=.24\textwidth]{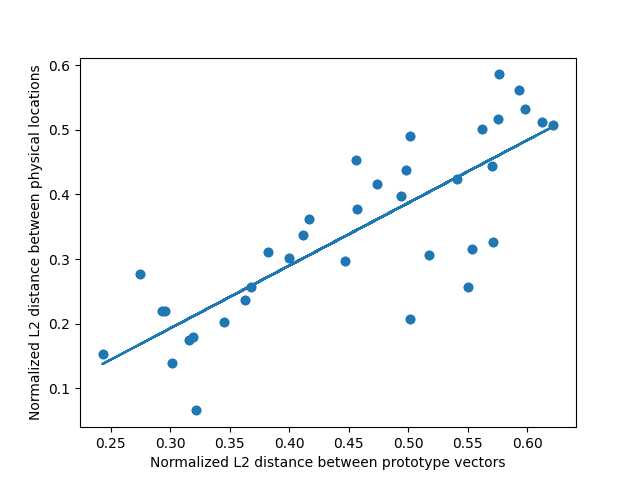}
    \includegraphics[width=.24\textwidth]{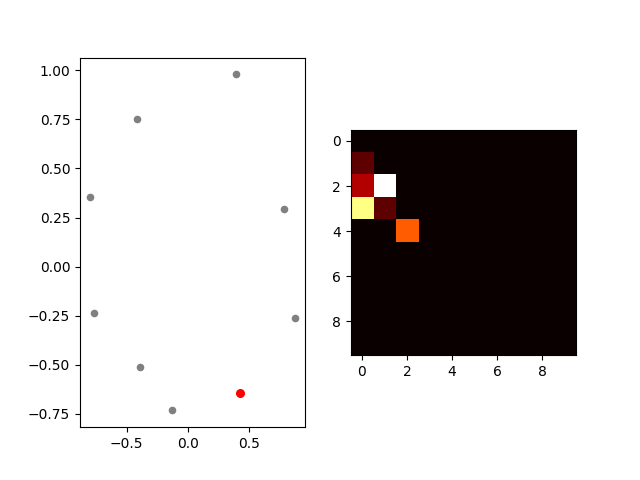}
    \includegraphics[width=.24\textwidth]{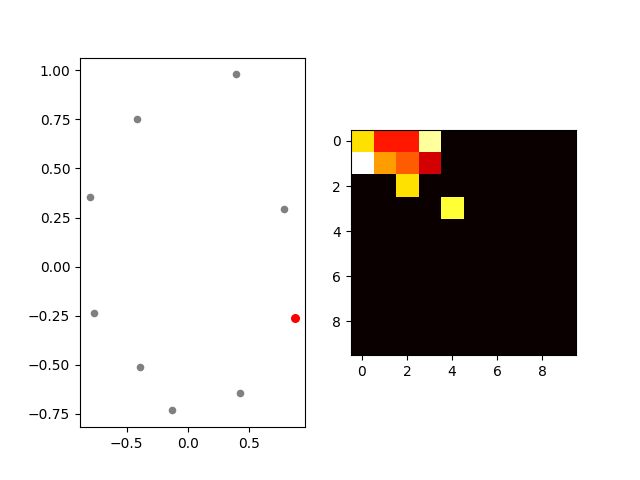}
    \caption{The distance between a pair of prototypes was highly correlated with the distance between the locations the prototypes referred to (left plot). A single prototype referred to a cluster of nearby locations (middle and right plots), with lighter colors denoting more frequent locations. Here, the two prototypes were both in the bottom right of the 2D PCA of the communication space and both referred to locations in the upper left of the grid.}
    \label{fig:corr_hc}
\end{figure*}

\begin{figure}
    \centering
    \includegraphics[width=.48\columnwidth]{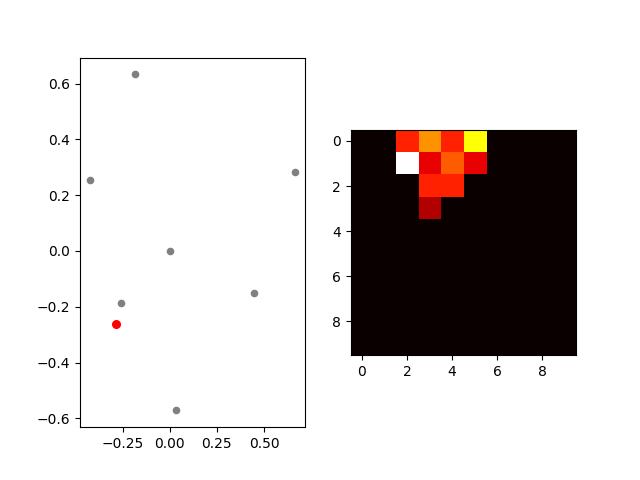}
    \includegraphics[width=.48\columnwidth]{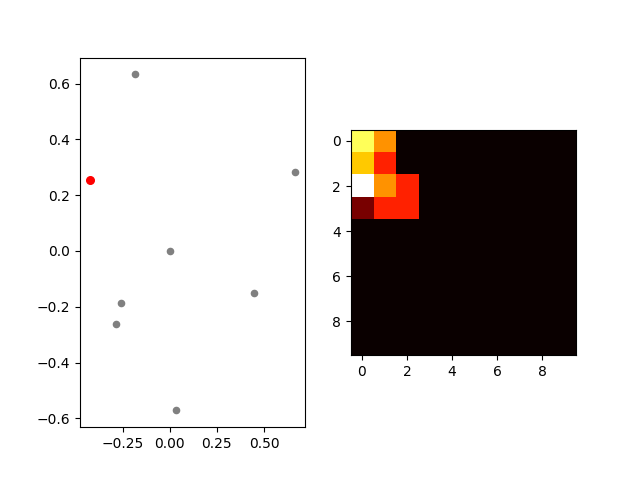}
    \caption{1-hot communication vectors show no correlation between prototype locations and environment locations.}
    \label{fig:1hot_hc}
\end{figure}

\section{Agent Interpretability} \label{agent_interpretability}
This section discusses the analysis of the properties that we observed in the agent training experiments and deem useful for learning  the prototypes and sparsity of communication.
Using domain knowledge of the task, we can deconstruct the prototypes to understand latent space communication. We performed a 2D Principal Component Analysis (PCA) of the prototypes for each environment and task based on the agents' observations while communicating their respective messages.

\subsection{Setup}
Our purpose of interpretability is to find a post-hoc analysis of the emergent message paradigm learned in self-play. We analyze the degree to which we can determine the white box meaning of learned messages.
We compare 1-hot encodings with prototype encodings.
We analyze the learned messages in the Medium Traffic Junction scenario as described in section~\ref{sec:experimental_setup} and the Parent and Lost Child scenario, which is a single predator and single prey subcase of Predator-Prey as described in section~\ref{sec:experimental_setup}.
Simply, or analysis aims to solve the follow question:
What are the properties of the different emergent communication policies that we foresee being most useful in HAT?

\subsection{Methodology}
From intuitive and mathematical perspectives, we analyze how agents convey observation information using prototype or one-hot messages. 
First, we visualize associations between prototypes and agent locations (a subset of observation information), as shown in the rightmost figures in Fig.~\ref{fig:corr_hc}. The eight gray dots and one red dot denote the 2D principal component analysis (PCA) of the 9 prototypes that the agent used.
The red dot denotes a particular prototype; the heatmap to the right of the PCA plot shows which agent locations were observed when the agent emitted the denoted prototype.
Intuitively, these heatmaps show the meaning of each prototype. Additionally, we analyze the relationship between the Euclidean distance between vectors and the information that they contain, e.g., in Parent and Lost Child the information is environment locations, in order to determine any correlation. Ideally, we want to see structured interpolation between information and messages in the latent space to prevent null messages, which contain ambiguous or no information.

\subsection{Results}

\subsubsection{Parent and Lost Child}

\begin{figure}[!t]
    \centering
    \includegraphics[width=.48\columnwidth]{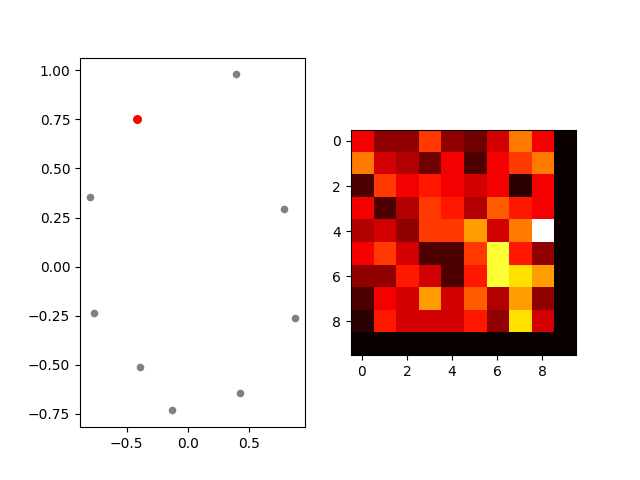}
        \includegraphics[width=.48\columnwidth]{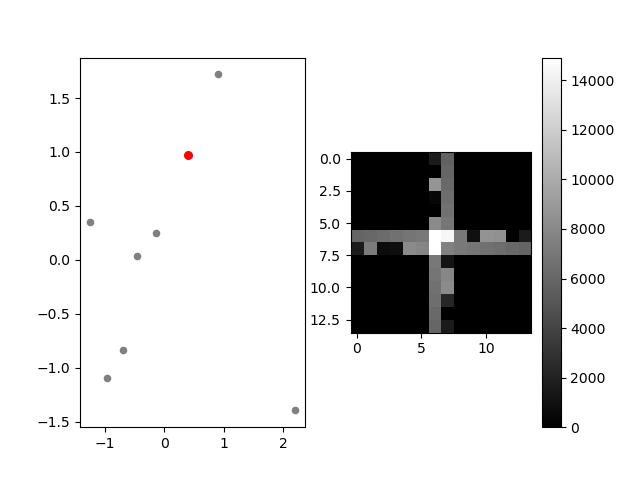}
    \caption{Recurrence with learned prototypes leads to null prototypes for communication. That is, it conveys no information to the other agent.  These communications are unnecessary and show that sparsity is feasible. Left: Human and Lost Child. Right: Traffic Junction.}
    \label{fig:proto_hc}
\end{figure}

\textbf{Property 1: Relational Observation Encodings.}
We found that the learned prototypes exhibited several desirable characteristics.
First, only 9 prototypes were used, indicating that the agents divided the $9 \times 9$ grid into coarser areas.
Second, each prototype referred to a distinct patch of the grid.
Lastly, and most interestingly, prototypes that were close together in communication space referred to nearby locations in the grid.
For example, the two visualized prototypes in Fig.~\ref{fig:corr_hc} are both in the bottom right of the communication space, and both refer to grid locations in the upper right.
We confirmed this anecdotal evidence by finding that the distance between prototypes and the distance between the locations the prototypes referred to was tightly correlated (with an $r^2 \approx 0.5$).

While one-hot based communication also successfully represented grid locations, as shown in Fig.~\ref{fig:1hot_hc}, it was fundamentally unable to exhibit a similar correlation as found between environment location and prototype location with prototypes.
All one-hot vectors are equally far apart by definition, so one cannot predict grid distances as a function of distance between communication vectors ($r^2 = 0$). 
Furthermore, given that all one-hot vectors are orthogonal, conducting PCA on the vectors does not provide any information on distance relations.

\subsubsection{Traffic Junction}

\begin{figure*}[!t]
    \centering
    \includegraphics[width=.24\textwidth]{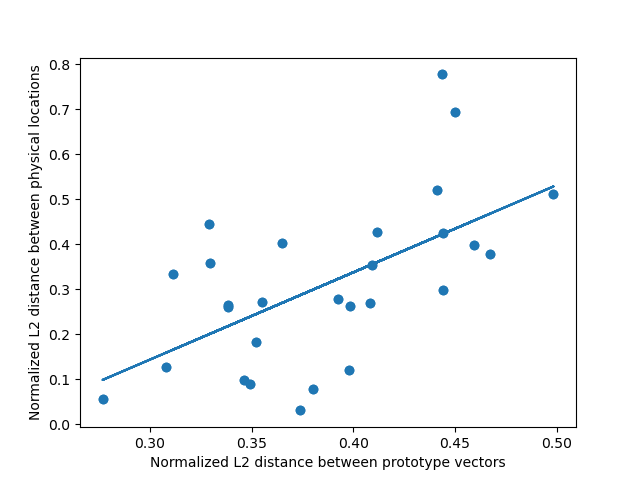}
    \includegraphics[width=.24\textwidth]{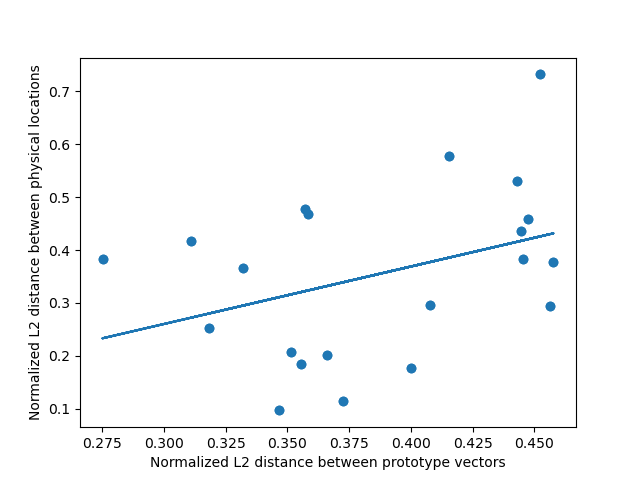}
    \includegraphics[width=.24\textwidth]{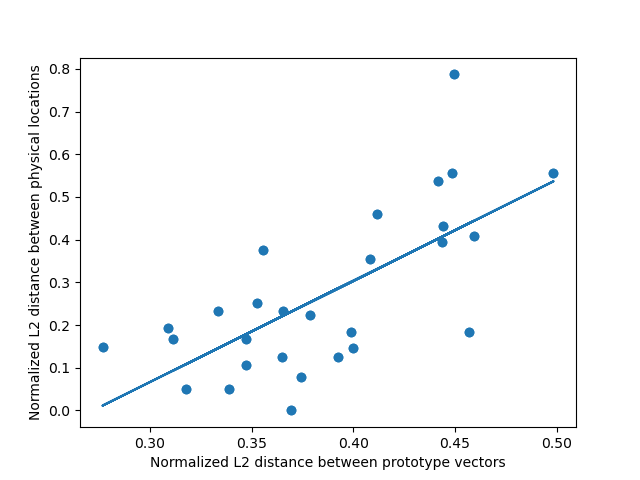}
    \caption{Above shows the correlation for prototypes in Traffic Junction with full communication (left), medium sparsity at 50\% communication (middle), and optimal sparsity at 30\% communication (right).}
    \label{fig:tj_corr}
\end{figure*}

\begin{figure*}
    \centering
        \includegraphics[width=.24\textwidth]{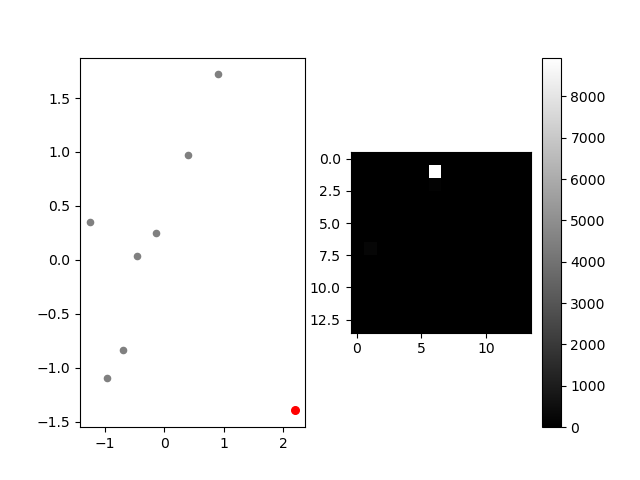}
        \includegraphics[width=.24\textwidth]{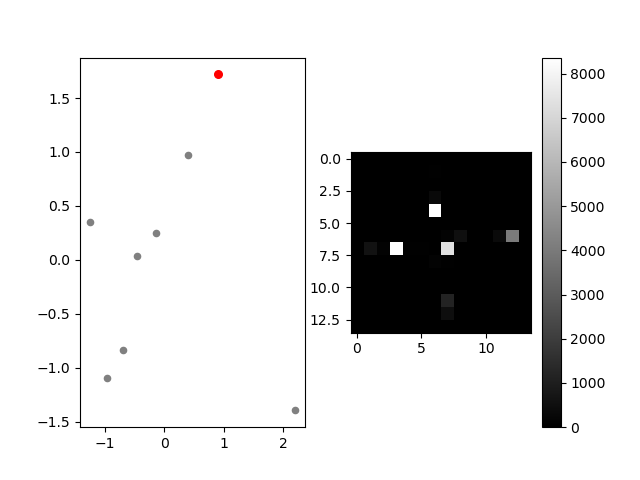}
        \includegraphics[width=.24\textwidth]{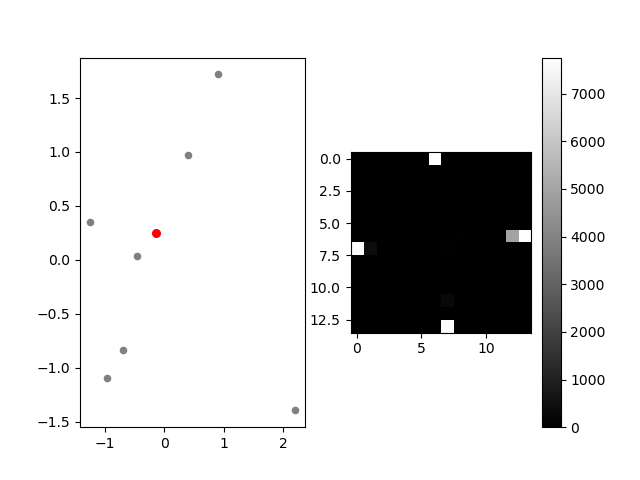}
    \caption{Above shows the Traffic Junction locations (right figures) represented by their corresponding prototype as shown in a principal component analysis (left figures).}
    \label{fig:tj_proto_loc}
    \vspace{-6mm}
\end{figure*}

\textbf{Property 2: Sparse Communication Through Information Content.}
Results from prototype-based communication vectors yield that they encode intrinsic information of the environment locations. 
Fig.~\ref{fig:tj_corr} shows the correlation of the Euclidean distance between each prototype vector to each other and the corresponding Euclidean distance in the environment. The slope for the full continuous communication, 50\% communication (mid-sparse), and 30\% communication (max sparsity) are 1.94, 1.09, and 2.36, respectively. The emergent prototypes tend to have multiple location mappings per distinct prototype. This results from the prototypes often representing information similarly per lane instead of uniquely for each environment location.

Since navigating the traffic junction ``blind'' is a complex task, communicating the location alone is insufficient information to avoid a collision. One may think that the agent's intent is also necessary to interpret. A car may communicate its location, but it is useful to know if they are going to accelerate (move to the next cell) or brake (stay in the same cell). However, the agents communicate the same prototype token regardless of the action that they take, which implies that the agents rely on recurrent and coordination information within their communications. This follows from homogeneous action policies among agents.

Recurrence and other coordination information play a large role for the agents in Traffic Junction. The agents tend to announce themselves when they enter the environment, but afterwards they communicate only at key points depending on the direction that they are traveling. Additionally, the agents repeat prototypes irrespective of the lane. See Fig.~\ref{fig:tj_proto_loc}. This means that agents send less information later in their trajectory, implying that agents can remember important details from previous messages.
In Fig.~\ref{fig:proto_hc}, the last figure shows that a null prototype is often used. This communication vector shows that, while the agents will still optimally traverse the junction, at least 30\% of all communications may be gated (prevented from being communicated).

\section{Human-Agent Teaming} \label{HAT}
This section discusses the setup and results of the human-agent teaming experiments. In previous sections, we introduced our multi-agent sparse-discrete emergent communication method and showed its effectiveness in agent self-play experiments. The next intuitive question is: would the communication method work when a human is swapped with one of the agents in the original problem setup? Although humans and MARL agents use entirely different communication systems, we deliberately design two human subject experiments with reasonable translations and approximations in order to evaluate our method in multiple human-agent teaming task scenarios. Specifically, two hypotheses are tested: 
    
    \begin{itemize}
    \item \textbf{H1}: Humans can learn to communicate in a human-agent team using the emergent prototypes from agent self-play.
    \item \textbf{H2}: Sparsity in communication reduces the cognitive workload on a human teammate such that they are able to perform better at the task.

\end{itemize}

\subsection{Human Experiment Design}

\begin{figure*}
    \centering
    \includegraphics[height=.24\textwidth]{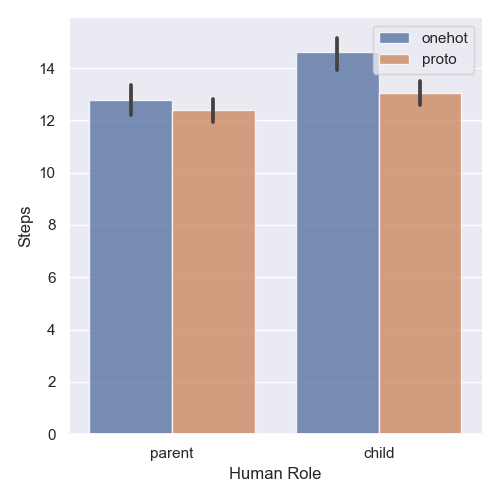}
    \includegraphics[height=.24\textwidth]{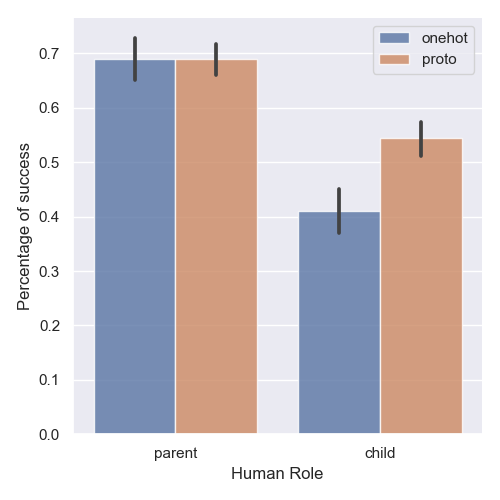}
    \includegraphics[height=.24\textwidth]{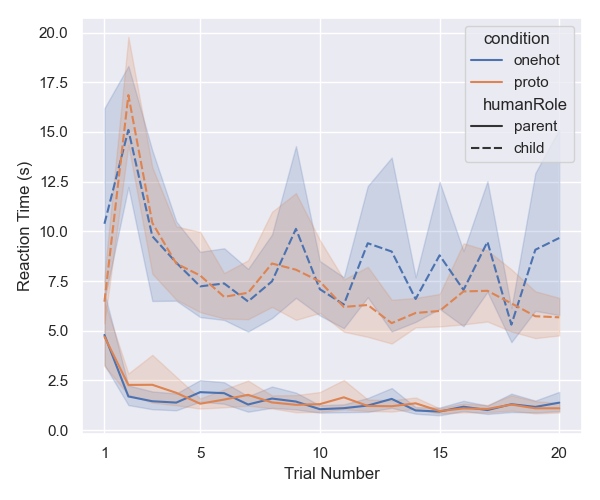}
    \caption{Learning curves of human-agent communication in the parent and lost child scenario. Left: Average steps taken per trial in one-hot and prototype conditions. Middle: Percentages of completed trials by roles and conditions. Right: Reaction time learning curves of human participants in the child (dashed lines) and parent roles (solid lines). Shaded areas indicate 95\% confidence intervals. }
    \label{fig:human_pp}
\end{figure*}

\begin{figure}
    \centering
    \includegraphics[width=\columnwidth]{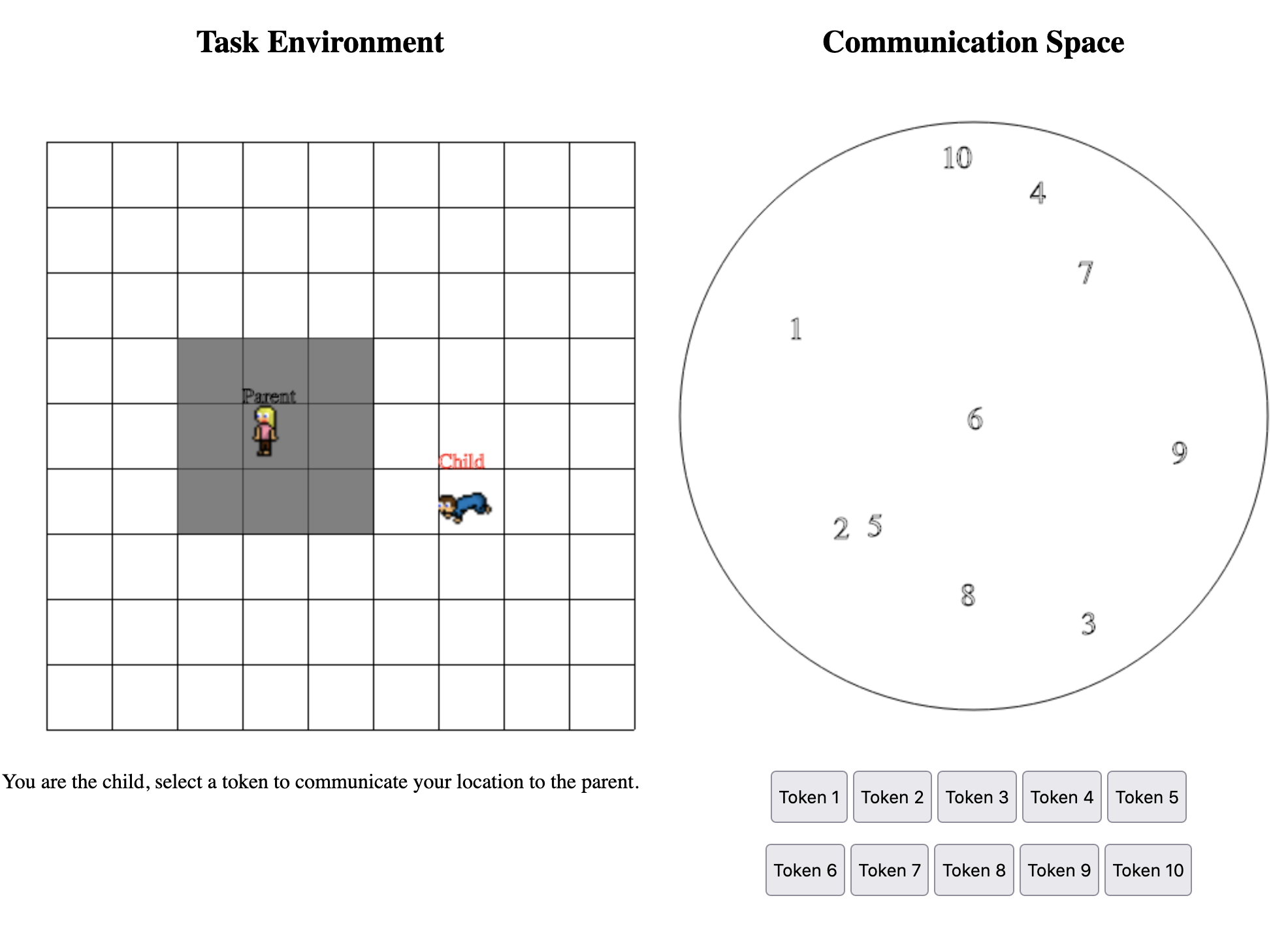}
    \caption{User interface of the human-agent communication experiment in the Parent and Lost Child scenario. The left panel displays the game environment state, including the locations and vision of the parent and child. The right panel consists of a communication space and selection buttons. Communication tokens are arranged in the 2D space based on PCA analysis to indicate potential semantic meanings. }
    \label{fig:pp_human}
\end{figure}

\subsubsection{Interpretability}
The core of \textbf{H1} lies in the interpretability of communication tokens learned by MARL agents. If the emergent ``language'' is interpretable, then the human can effectively develop a mental model and use it to collaborate with agent teammates. Participants in the first human-agent teaming experiment teamed with MARL agents to solve the Parent and Lost Child task. Half of the participants use prototype-based communication, and the other half use one-hot communication. The first experiment aims to test whether humans can learn the mapping relationship between tokens and child locations better than rote memorization and use it in a human-agent team task. The prototype vectors are translated into communication tokens and plotted on a 2D space using PCA as described in section~\ref{agent_interpretability}. For 1-hot vectors, we use the same spatial arrangement with prototype tokens but randomize the token mapping between subjects. By doing so, we eliminate the confounding effect of spatial structure in human learning and can better focus on the core comparison between prototype-based and 1-hot communication. 

\subsubsection{Sparsity}
Human-agent communication is usually limited by the divergent cognitive capacity of humans and agents in information processing. That is, humans are good at spatial, heuristic, and analogical reasoning while autonomous agents process high-frequency information continuously~\cite{fan2010modeling}. \textbf{H2} assesses if introducing sparse messaging helps humans process the information sent by agents and lower the overall cognitive workload. We evaluate this hypothesis in the Traffic Junction experiment, where a human participant is swapped with one of the cars in lane. Other cars are MARL agents trained with a budget only to send messages when necessary, as described in section~\ref{agent_training}. A ``ghost'' agent processes incoming messages to facilitate human understanding, which is translated into a graphic user interface based on the prior PCA analysis in section~\ref{agent_interpretability}. Since the analyzed communication method does not create an injection with environment locations, the car's location is represented as a probability distribution. A caution triangle visualizes the likelihood of a cell being occupied by other agents on the user interface. Messages from the human are handled by a ``ghost'' agent, which acts as an interface between the human and messages from/to artificial agents. Previous literature has shown this method to be effective in studying communication understanding in HATs~\cite{li2015communication}. By implementing the experiment in this way, we directly present the meaning of the prototype communication to the human to speed up the learning process. Thus, the experiment can better focus on the influence of sparsity on human understanding of agent communication.

\begin{figure*}[!t]
    \centering
    
    \includegraphics[height=.24\textwidth]{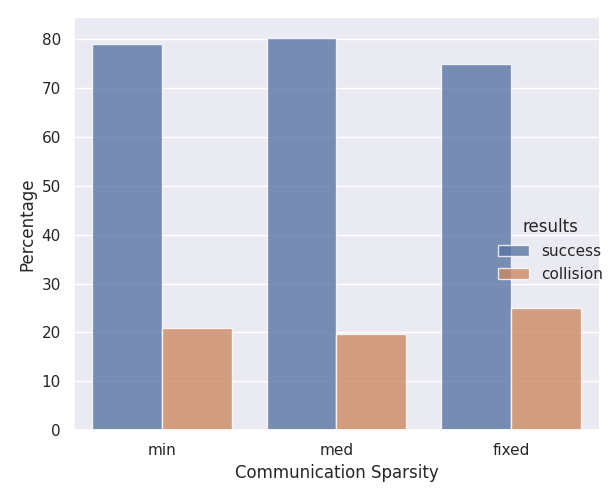}
    \includegraphics[height=.24\textwidth]{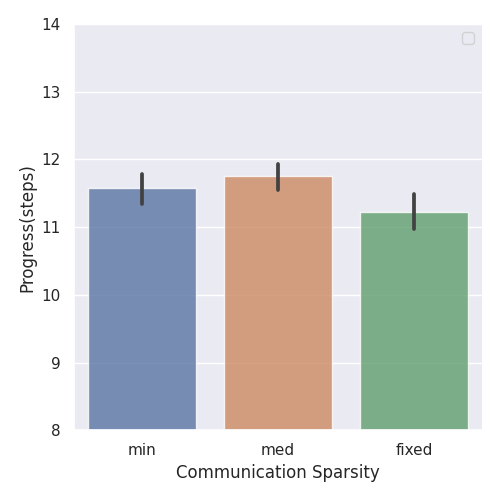}
    \includegraphics[height=.24\textwidth]{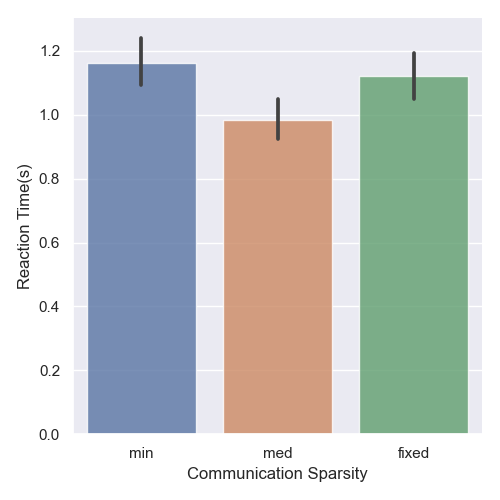}
    \caption{Learning curves of human-agent communication in Traffic Junction. Left: Percentages of success in trials with three communication sparsity conditions. Middle: Average progress made per trial. Right: Average reaction time.  Error bars indicate 95\% confidence intervals. }
    \label{fig:human_tj}
\end{figure*}

\begin{figure}
    \centering
    \includegraphics[width=.4\columnwidth]{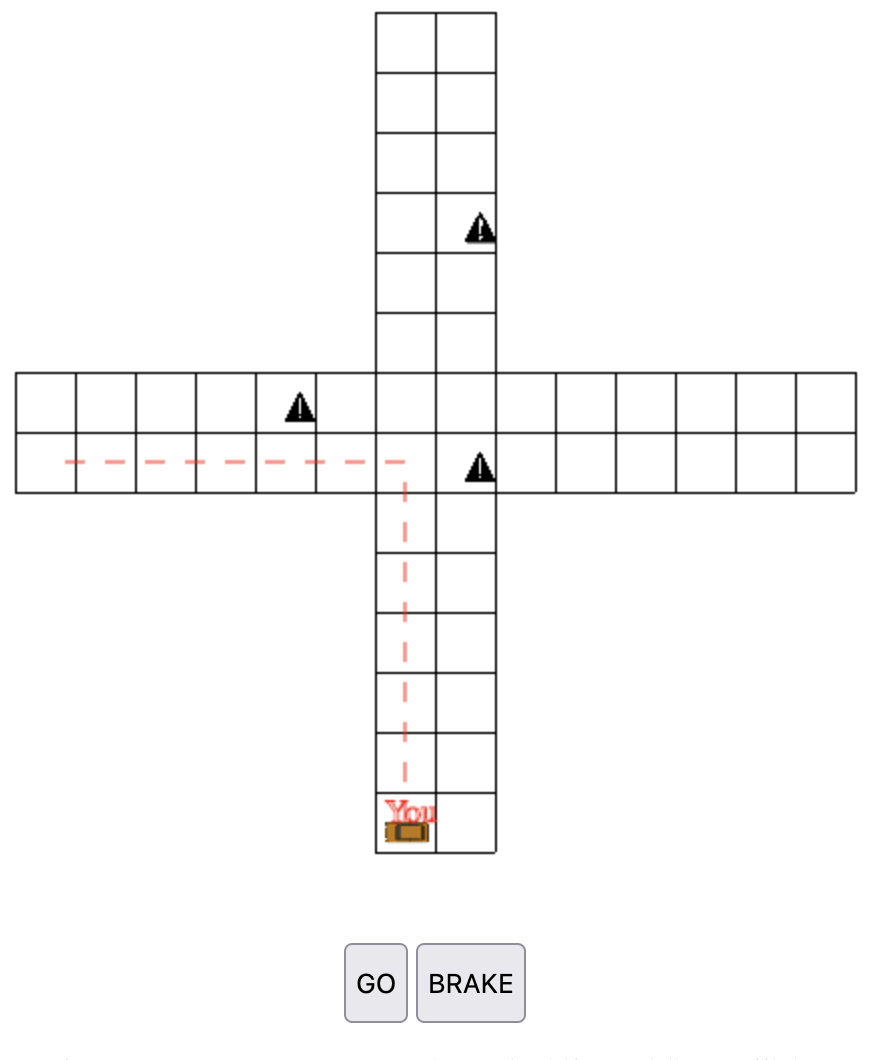}
    \caption{The human interface for Traffic Junction is shown above. The participants' task is to move the yellow car along the designated path (red dashed line) using two actions: go or brake. Black caution triangles refer to potential locations of other cars as revealed by agent communication. }
    \label{fig:tj_human}
\end{figure}

\subsection{Methodology}

\subsubsection{Parent and Lost Child}
Each participant is teamed up with an AI agent in the Parent and Lost Child environment. The humans complete 20 trials in each parent and child role, with corresponding communication tasks. We counterbalance the sequence of two experimental sessions between subjects. At the beginning of the trial, the parent and child spawn at random locations. A trial ends when the parent reaches the child or exceeds the maximum step limitation (20). The human-agent team performance is measured by 1) the number of steps the parent takes to find the child, 2) the percentage of trials in which the parent successfully found the child, and 3) the time it takes the human to make decisions.

The task interface, shown in Fig.~\ref{fig:pp_human}, consists of two panels: the task environment and the communication panel. The task environment displays the true game state, the parent's vision radius, and the parent's trajectories. Interface visibility is subject to change according to the human's role. The communication panel presents numbered tokens for the child to send its location. The number and arrangement of tokens are determined by the 2D PCA analysis from section~\ref{agent_interpretability}. Once the child selects a token, it is highlighted on the panel.

When the human is playing the parent role, one fixed token is highlighted on the communication panel throughout the trial to signal to the human what the child agent is communicating. The human then must select a movement action to move using arrow keys on their keyboard. The participants' task as the parent is to learn to understand the meaning of the communication tokens in order to find the child in the least number of steps. When the human is in the child role, she may select only one communication token to send to the parent agent per trial but can use a different token in different trials. The human then views a replay of the entire trajectory of the parent agent as feedback to determine if the communication is helpful. Similarly, participants in the child role do not know the semantic meanings of each communication token at the beginning. Therefore, they must learn to develop the mapping between tokens and locations via repeated interactions.

\subsubsection{Traffic Junction}
In order to evaluate the effect of sparsity, human participants were recruited to team up with multiple agents in Traffic Junction. Fig.~\ref{fig:tj_human} shows the user interface of the Traffic Junction experiment. Humans act as one of the cars traversing through the junction. The goal of this task is to traverse through one's lane as far as possible without collision. All cars, including both the human and the agents, are 'blind' and have predetermined goals. The human car will enter the environment while between 5 and 9 agents are currently navigating the junction. Each trial lasts for 40 steps. The human can choose whether to accelerate or brake on the designated path to navigate the intersection without collision before the end of the trial.

\subsection{Results}

\subsubsection{Human and Lost Child}
106 participants were recruited from Amazon MTurk and Prolific, 30 were removed due to an incomplete session or failure to pass the attention check. Participants were randomly divided into two conditions, receiving either one-hot communication or prototype communication tokens. The average number of steps taken per trial is  $12.74$ in the prototype condition and $13.64$ in the one-hot condition. This measurement quantifies the collaboration outcome of the human-agent team. The fewer steps taken, the better the team's performance.  
We conducted a mixed-ANOVA in which the human role is the within-subject variable and the experimental condition is the between-subject variable. 
Results show that the main effects of the condition are significant ($F(1,74) = 5.95,\ p = .017$), indicating the proposed prototype communication indeed leads to better team performance as compared to the one-hot baseline. In addition, the main effect of the role and interaction effect between the condition and role are both significant $(p < .05)$ for each one. Paired t-tests show that prototype communication significantly outperforms one-hot baseline in the human child condition (13.09 vs. 14.54, $ t(52.4) = 3.43,\ p = .001,\ cohen\ d = .83$), but not in the human parent condition. The other measurement of human-agent team performance is the completion rate shown in Fig.~\ref{fig:human_pp}-middle. Chi-squared analysis shows similar patterns as above: prototype communication leads to significantly better team performance, and this difference is mainly due to the human child condition.

Another research question is how can one reveal a human's learning process of a communication language generated by reinforcement learning agents. To answer this question, we measure the reaction time of humans to decide what action to take or what communication token to send as an indicator of participants' cognitive workload. As shown in Fig.~\ref{fig:human_pp}-right, the required reaction time decreases along the course of interaction between the human and agent teammates. Trial number is negatively correlated with reaction time in both child ($r = -.17,\ p < .001$) and parent ($r = -.24,\ p < .001$) roles. Specifically, this learning effect is more substantial when the human child was using prototype-based communication ($r = -.24,\ p < .001$) as compared to one-hot communication ($r = -.08,\ p = .045$). While the learning effect of team performance is not observed directly, 
participants actually learn the communication language during the interaction with reinforcement learning agents and complete the task faster. Our proposed prototype communication can speed up this learning process by allowing the human to bear a lower cognitive load and select the correct communication token more quickly in the child role.

In summary, \textbf{H1} confirmed that 1) prototype communication is interpretable to humans and leads to better HAT performance, and 2) prototype communication speeds up the learning process of humans by lowering the cognitive load. Our proposed method is especially effective when the human sends messages as the child. A possible explanation is the different levels of initiative between human (exploratory) and agent (exploitative) parent's searching strategies. When receiving an invalid communication, unlike the agent, the human will continue exploration.

\subsubsection{Traffic Junction: Evaluating Sparsity}
142 human participants were recruited from Amazon MTurk and Prolific, but 16 were removed due to incomplete sessions. Participants were divided into three conditions, either use prototype communication with fixed non-sparse $b=1$, 
medium sparsity $b=0.5$, or minimum sparsity $b=b^*=0.3$. This condition division is based on agent training results, in which $b^*$ was shown to be the minimum frequency without sacrificing performance.
Each participant completed 20 trials of the experiment. Team performance in the Traffic Junction task is measured by the percentage of trials in which a human's car successfully arrived at its destination without collision. The average success rate for the three conditions is min: 79.1\%, med: 80.1\%, fixed: 74.6\% (as shown in Fig.~\ref{fig:human_tj}-left). A chi-squared test shows that the relationship between task success and communication sparsity is significant ($\chi^2 (2,N=2408) = 7.28,\ p = .026$). Human participants who received sparse communication (i.e., min and med conditions) from agents are more likely to succeed in passing through the junction. 

Recording how far each human managed to proceed before collision is also a good measure of a human's task progress. Fig.~\ref{fig:human_tj}-middle shows the average progress 
Similar to the above analysis, one-way ANOVA shows a significant difference between conditions ($F(2,2405) = 5.27,\ p = .005$). T-tests indicate that both min ($11.57$) and med ($11.74$) conditions outperform the fixed baseline ($11.23$) in making more progress ($p <.05$) for each condition.

To reveal the reasons of performance improvement brought by sparse communication, we plot the average reaction time of participants in three conditions in Fig.~\ref{fig:human_tj}-right. ANOVA analysis shows a significant difference in reaction time between conditions ($F(2,2405) = 7.49,\ p = .001$). Paired t-tests show that participants in the medium sparsity communication condition had a significantly shorter reaction time ($0.98$s) than those in either minimum ($1.16$s) and full communication ($1.12$s) conditions. Considering reaction time as an indicator of human cognitive workload, we found an inverted 'U' shape relationship between communication sparsity and human cognitive load. Our proposed sparse method with appropriate configuration was shown to reduce human cognitive load by sending fewer communication messages; that is, messages are sent only at necessary moments, confirming \textbf{H2}.

\section{Conclusion} \label{conclusion}

Recent work in MARL has aimed to develop sparse-discrete communication paradigms.
In this work, we have analyzed a sparse-discrete method to determine its interpretability and performance when used in human-agent teaming. 
Through the Enforcers we can train sparse-discrete models to perform competitively with unconstrained alternatives.
 The results show that the intent of the communication can be trained to correlate with human-understandable observations of information necessary to complete tasks.

We conducted two human-agent teaming experiments to explore the relationship between the human learning effect and various properties of agent communication. 
In the first experiment, we evaluated whether humans are able to learn the communication “language” generated by reinforcement learning agents. We compared our proposed interpretable prototype-based communication against a one-hot encoding communication baseline. Results validate the superiority of prototype-based communication in better overall team performance and a faster learning effect over the baseline, confirming our proposed method's interpretability. Based on the analysis in section~\ref{agent_interpretability}, we attribute these results to the correlation relationship between tokens' locations in the communication space and referenced child locations in the task environment. 
Prototype-based messages have a structured latent space, allowing for learning ease. Our results verify this hypothesis since there is a steeper learning curve of reaction time to decide what communication token to send.

The results of the traffic junction experiment support our hypothesis about communication sparsity in human-agent teaming with multiple agents. Both minimum and medium sparse communication enable better team performance than the full communication baseline. Measurements of reaction time as a proxy for human cognitive workload~\cite{van2018effects} provided further corroborating evidence. 
Reaction times followed an inverted 'U' shape, in which medium communication sparsity led to lower cognitive loads than minimum or maximum frequency; one explanation for this trend is that overly-frequent communication overloads humans, while overly-sparse communication is hard to recall.
These findings align with previous literature~\cite{marlow2018does} and imply that introducing an appropriate communication frequency budget is essential in supporting human-agent interaction.
As emphasized by our agent self-play and HAT experiments, adapting communication to many sparse budgets is critical for effective human-agent teaming.
An additional interesting discovery was that an adaptive function might be necessary to find the appropriate communication sparsity for each individual human teammate or population of humans. We will pursue this research avenue in future work.

\section*{Acknowledgments}
This work was partially funded by ARL award W911NF1920146, DARPA award HR001120C0036 and AFOSR award FAA95501810251.

\bibliographystyle{IEEEtran}
\bibliography{bib}

\begin{thebibliography}{10}
\providecommand{\url}[1]{#1}
\csname url@samestyle\endcsname
\providecommand{\newblock}{\relax}
\providecommand{\bibinfo}[2]{#2}
\providecommand{\BIBentrySTDinterwordspacing}{\spaceskip=0pt\relax}
\providecommand{\BIBentryALTinterwordstretchfactor}{4}
\providecommand{\BIBentryALTinterwordspacing}{\spaceskip=\fontdimen2\font plus
\BIBentryALTinterwordstretchfactor\fontdimen3\font minus
  \fontdimen4\font\relax}
\providecommand{\BIBforeignlanguage}[2]{{%
\expandafter\ifx\csname l@#1\endcsname\relax
\typeout{** WARNING: IEEEtran.bst: No hyphenation pattern has been}%
\typeout{** loaded for the language `#1'. Using the pattern for}%
\typeout{** the default language instead.}%
\else
\language=\csname l@#1\endcsname
\fi
#2}}
\providecommand{\BIBdecl}{\relax}
\BIBdecl

\bibitem{gronauer2021multi}
S.~Gronauer and K.~Diepold, ``Multi-agent deep reinforcement learning: a
  survey,'' \emph{Artificial Intelligence Review}, pp. 1--49, 2021.

\bibitem{zhang2021multi}
K.~Zhang, Z.~Yang, and T.~Ba{\c{s}}ar, ``Multi-agent reinforcement learning: A
  selective overview of theories and algorithms,'' \emph{Handbook of
  Reinforcement Learning and Control}, pp. 321--384, 2021.

\bibitem{starcraft}
\BIBentryALTinterwordspacing
P.~Peng, Q.~Yuan, Y.~Wen, Y.~Yang, Z.~Tang, H.~Long, and J.~Wang, ``Multiagent
  bidirectionally-coordinated nets for learning to play starcraft combat
  games,'' \emph{CoRR}, vol. abs/1703.10069, 2017. [Online]. Available:
  \url{http://arxiv.org/abs/1703.10069}
\BIBentrySTDinterwordspacing

\bibitem{dota2}
C.~Berner, G.~Brockman, B.~Chan, V.~Cheung, P.~D{\k{e}}biak, C.~Dennison,
  D.~Farhi, Q.~Fischer, S.~Hashme, C.~Hesse \emph{et~al.}, ``Dota 2 with large
  scale deep reinforcement learning,'' \emph{arXiv preprint arXiv:1912.06680},
  2019.

\bibitem{siu2021evaluation}
H.~C. Siu, J.~Pe{\~n}a, E.~Chen, Y.~Zhou, V.~Lopez, K.~Palko, K.~Chang, and
  R.~Allen, ``Evaluation of human-ai teams for learned and rule-based agents in
  hanabi,'' \emph{Advances in Neural Information Processing Systems}, vol.~34,
  2021.

\bibitem{carroll2019utility}
M.~Carroll, R.~Shah, M.~K. Ho, T.~Griffiths, S.~Seshia, P.~Abbeel, and
  A.~Dragan, ``On the utility of learning about humans for human-ai
  coordination,'' \emph{Advances in Neural Information Processing Systems},
  vol.~32, pp. 5174--5185, 2019.

\bibitem{commnet}
S.~Sukhbaatar, R.~Fergus \emph{et~al.}, ``Learning multiagent communication
  with backpropagation,'' \emph{Advances in neural information processing
  systems}, vol.~29, pp. 2244--2252, 2016.

\bibitem{ic3net}
A.~Singh, T.~Jain, and S.~Sukhbaatar, ``Learning when to communicate at scale
  in multiagent cooperative and competitive tasks,'' in \emph{International
  Conference on Learning Representations}, 2018.

\bibitem{andreas2017translating}
J.~Andreas, A.~Dragan, and D.~Klein, ``Translating neuralese,'' in
  \emph{Proceedings of the 55th Annual Meeting of the Association for
  Computational Linguistics (Volume 1: Long Papers)}, 2017, pp. 232--242.

\bibitem{nikolaidis2013}
S.~Nikolaidis and J.~Shah, ``Human-robot cross-training: Computational
  formulation, modeling and evaluation of a human team training strategy,'' in
  \emph{2013 8th ACM/IEEE International Conference on Human-Robot Interaction
  (HRI)}, 2013, pp. 33--40.

\bibitem{soltani2016}
A.~Soltani, P.~Khorsand, C.~Guo, and J.~Liu, ``Neural substrates of cognitive
  biases during probabilistic inference,'' \emph{Nature Communications},
  vol.~7, no.~1, p. e1000858, 2016.

\bibitem{kliegr2021}
\BIBentryALTinterwordspacing
T.~Kliegr, Štěpán Bahník, and J.~Fürnkranz, ``A review of possible effects
  of cognitive biases on interpretation of rule-based machine learning
  models,'' \emph{Artificial Intelligence}, vol. 295, p. 103458, 2021.
  [Online]. Available:
  \url{https://www.sciencedirect.com/science/article/pii/S0004370221000096}
\BIBentrySTDinterwordspacing

\bibitem{silver2017}
\BIBentryALTinterwordspacing
D.~Silver, J.~Schrittwieser, K.~Simonyan, I.~Antonoglou, A.~Huang, A.~Guez,
  T.~Hubert, L.~Baker, M.~Lai, A.~Bolton, Y.~Chen, T.~Lillicrap, F.~Hui,
  L.~Sifre, G.~van~den Driessche, T.~Graepel, and D.~Hassabis, ``Mastering the
  game of go without human knowledge,'' \emph{Nature}, vol. 550, pp. 354--,
  Oct. 2017. [Online]. Available: \url{http://dx.doi.org/10.1038/nature24270}
\BIBentrySTDinterwordspacing

\bibitem{chan2017}
D.~Chan, ``The ai that has nothing to learn from humans,'' \emph{The Atlantic},
  vol.~7, no.~1, p. e1000858, 2017.

\bibitem{vinyals2019}
O.~Vinyals, I.~Babuschkin, W.~M. Czarnecki, M.~Mathieu, A.~Dudzik, J.~Chung,
  D.~H. Choi, R.~Powell, T.~Ewalds, P.~Georgiev, J.~Oh, D.~Horgan, M.~Kroiss,
  I.~Danihelka, A.~Huang, L.~Sifre, T.~Cai, J.~P. Agapiou, M.~Jaderberg, A.~S.
  Vezhnevets, R.~Leblond, T.~Pohlen, V.~Dalibard, D.~Budden, Y.~Sulsky,
  J.~Molloy, T.~L. Paine, C.~Gulcehre, Z.~Wang, T.~Pfaff, Y.~Wu, R.~Ring,
  D.~Yogatama, D.~W{\"u}nsch, K.~McKinney, O.~Smith, T.~Schaul, T.~P.
  Lillicrap, K.~Kavukcuoglu, D.~Hassabis, C.~Apps, and D.~Silver, ``Grandmaster
  level in starcraft ii using multi-agent reinforcement learning,''
  \emph{Nature}, pp. 1--5, 2019.

\bibitem{iyer2018transparency}
R.~Iyer, Y.~Li, H.~Li, M.~Lewis, R.~Sundar, and K.~Sycara, ``Transparency and
  explanation in deep reinforcement learning neural networks,'' in
  \emph{Proceedings of the 2018 AAAI/ACM Conference on AI, Ethics, and
  Society}, 2018, pp. 144--150.

\bibitem{anderson2019}
\BIBentryALTinterwordspacing
A.~Anderson, J.~Dodge, A.~Sadarangani, Z.~Juozapaitis, E.~Newman, J.~Irvine,
  S.~Chattopadhyay, A.~Fern, and M.~Burnett, ``Explaining reinforcement
  learning to mere mortals: An empirical study,'' in \emph{Proceedings of the
  Twenty-Eighth International Joint Conference on Artificial Intelligence,
  {IJCAI-19}}.\hskip 1em plus 0.5em minus 0.4em\relax International Joint
  Conferences on Artificial Intelligence Organization, 7 2019, pp. 1328--1334.
  [Online]. Available: \url{https://doi.org/10.24963/ijcai.2019/184}
\BIBentrySTDinterwordspacing

\bibitem{marathe2018bidirectional}
A.~R. Marathe, K.~E. Schaefer, A.~W. Evans, and J.~S. Metcalfe, ``Bidirectional
  communication for effective human-agent teaming,'' in \emph{International
  Conference on Virtual, Augmented and Mixed Reality}.\hskip 1em plus 0.5em
  minus 0.4em\relax Springer, 2018, pp. 338--350.

\bibitem{lake2019human}
B.~M. Lake, T.~Linzen, and M.~Baroni, ``Human few-shot learning of
  compositional instructions,'' \emph{arXiv preprint arXiv:1901.04587}, 2019.

\bibitem{li2021learning}
S.~Li, Y.~Zhou, R.~Allen, and M.~J. Kochenderfer, ``Learning emergent discrete
  message communication for cooperative reinforcement learning,'' \emph{arXiv
  preprint arXiv:2102.12550}, 2021.

\bibitem{discreteComm}
M.~Tucker, H.~Li, S.~Agrawal, D.~Hughes, K.~Sycara, M.~Lewis, and J.~A. Shah,
  ``Emergent discrete communication in semantic spaces,'' \emph{Advances in
  Neural Information Processing Systems}, vol.~34, 2021.

\bibitem{marlow2018does}
S.~L. Marlow, C.~N. Lacerenza, J.~Paoletti, C.~S. Burke, and E.~Salas, ``Does
  team communication represent a one-size-fits-all approach?: A meta-analysis
  of team communication and performance,'' \emph{Organizational behavior and
  human decision processes}, vol. 144, pp. 145--170, 2018.

\bibitem{GreenSwets66}
D.~M. Green and J.~A. Swets, \emph{Signal Detection Theory and
  Psychophysics}.\hskip 1em plus 0.5em minus 0.4em\relax New York: Wiley, 1966.

\bibitem{van2005cognitive}
J.~J. Van~Merrienboer and J.~Sweller, ``Cognitive load theory and complex
  learning: Recent developments and future directions,'' \emph{Educational
  psychology review}, vol.~17, no.~2, pp. 147--177, 2005.

\bibitem{wang2020learning}
R.~Wang, X.~He, R.~Yu, W.~Qiu, B.~An, and Z.~Rabinovich, ``Learning efficient
  multi-agent communication: An information bottleneck approach,'' in
  \emph{International Conference on Machine Learning}.\hskip 1em plus 0.5em
  minus 0.4em\relax PMLR, 2020, pp. 9908--9918.

\bibitem{agrawal2021learning}
S.~Agrawal, ``Learning to imitate, adapt and communicate,'' Master's thesis,
  Carnegie Mellon University, 2021.

\bibitem{foerster2016learning}
J.~N. Foerster, Y.~M. Assael, N.~de~Freitas, and S.~Whiteson, ``Learning to
  communicate with deep multi-agent reinforcement learning,'' in
  \emph{Proceedings of the 30th International Conference on Neural Information
  Processing Systems}, 2016, pp. 2145--2153.

\bibitem{mao2020learning}
H.~Mao, Z.~Zhang, Z.~Xiao, Z.~Gong, and Y.~Ni, ``Learning agent communication
  under limited bandwidth by message pruning,'' in \emph{Proceedings of the
  AAAI Conference on Artificial Intelligence}, vol.~34, no.~04, 2020, pp.
  5142--5149.

\bibitem{vijay2021minimizing}
V.~K. Vijay, H.~Sheikh, S.~Majumdar, and M.~Phielipp, ``Minimizing
  communication while maximizing performance in multi-agent reinforcement
  learning,'' \emph{arXiv preprint arXiv:2106.08482}, 2021.

\bibitem{tarmac}
A.~Das, T.~Gervet, J.~Romoff, D.~Batra, D.~Parikh, M.~Rabbat, and J.~Pineau,
  ``Tarmac: Targeted multi-agent communication,'' in \emph{International
  Conference on Machine Learning}.\hskip 1em plus 0.5em minus 0.4em\relax PMLR,
  2019, pp. 1538--1546.

\bibitem{graphMA}
A.~Agarwal, S.~Kumar, K.~Sycara, and M.~Lewis, ``Learning transferable
  cooperative behavior in multi-agent teams,'' in \emph{Proceedings of the 19th
  International Conference on Autonomous Agents and MultiAgent Systems}, 2020,
  pp. 1741--1743.

\bibitem{goyal2020variational}
A.~Goyal, Y.~Bengio, M.~Botvinick, and S.~Levine, ``The variational bandwidth
  bottleneck: Stochastic evaluation on an information budget,'' \emph{arXiv
  preprint arXiv:2004.11935}, 2020.

\bibitem{kim2019learning}
D.~Kim, S.~Moon, D.~Hostallero, W.~J. Kang, T.~Lee, K.~Son, and Y.~Yi,
  ``Learning to schedule communication in multi-agent reinforcement learning,''
  in \emph{ICLR 2019: International Conference on Representation
  Learning}.\hskip 1em plus 0.5em minus 0.4em\relax International Conference on
  Representation Learning, 2019.

\bibitem{rashid2018qmix}
T.~Rashid, M.~Samvelyan, C.~Schroeder, G.~Farquhar, J.~Foerster, and
  S.~Whiteson, ``Qmix: Monotonic value function factorisation for deep
  multi-agent reinforcement learning,'' in \emph{International Conference on
  Machine Learning}.\hskip 1em plus 0.5em minus 0.4em\relax PMLR, 2018, pp.
  4295--4304.

\bibitem{seraj2022learning}
E.~Seraj, Z.~Wang, R.~Paleja, D.~Martin, M.~Sklar, A.~Patel, and M.~Gombolay,
  ``Learning efficient diverse communication for cooperative heterogeneous
  teaming,'' in \emph{Proceedings of the 21st International Conference on
  Autonomous Agents and Multiagent Systems}, 2022, pp. 1173--1182.

\bibitem{freed2020sparse}
B.~Freed, R.~James, G.~Sartoretti, and H.~Choset, ``Sparse discrete
  communication learning for multi-agent cooperation through backpropagation,''
  in \emph{2020 IEEE/RSJ International Conference on Intelligent Robots and
  Systems (IROS)}.\hskip 1em plus 0.5em minus 0.4em\relax IEEE, 2020, pp.
  7993--7998.

\bibitem{lowe2017multi}
R.~Lowe, Y.~Wu, A.~Tamar, J.~Harb, P.~Abbeel, and I.~Mordatch, ``Multi-agent
  actor-critic for mixed cooperative-competitive environments,'' in
  \emph{Proceedings of the 31st International Conference on Neural Information
  Processing Systems}, 2017, pp. 6382--6393.

\bibitem{mordatch2018emergence}
I.~Mordatch and P.~Abbeel, ``Emergence of grounded compositional language in
  multi-agent populations,'' in \emph{Thirty-second AAAI conference on
  artificial intelligence}, 2018.

\bibitem{peterson2017adapting}
J.~C. Peterson, J.~T. Abbott, and T.~L. Griffiths, ``Adapting deep network
  features to capture psychological representations: An abridged report.'' in
  \emph{IJCAI}, 2017, pp. 4934--4938.

\bibitem{visdial}
\BIBentryALTinterwordspacing
A.~Das, S.~Kottur, K.~Gupta, A.~Singh, D.~Yadav, J.~M.~F. Moura, D.~Parikh, and
  D.~Batra, ``Visual dialog,'' \emph{CoRR}, vol. abs/1611.08669, 2016.
  [Online]. Available: \url{http://arxiv.org/abs/1611.08669}
\BIBentrySTDinterwordspacing

\bibitem{agarwalvisdial2019}
A.~Agarwal, G.~Swaminathan, V.~Sharma, and K.~Sycara, ``Community
  regularization of visually-grounded dialog,'' in \emph{Proceedings of the
  2019 Conference on Autonomous Agents and Multiagent Systems (AAMAS)}, 2019.

\bibitem{lazaridou2016multi}
A.~Lazaridou, A.~Peysakhovich, and M.~Baroni, ``Multi-agent cooperation and the
  emergence of (natural) language,'' \emph{arXiv preprint arXiv:1612.07182},
  2016.

\bibitem{tse2007schemas}
D.~Tse, R.~F. Langston, M.~Kakeyama, I.~Bethus, P.~A. Spooner, E.~R. Wood,
  M.~P. Witter, and R.~G. Morris, ``Schemas and memory consolidation,''
  \emph{Science}, vol. 316, no. 5821, pp. 76--82, 2007.

\bibitem{li2021individualized}
H.~Li, T.~Ni, S.~Agrawal, F.~Jia, S.~Raja, Y.~Gui, D.~Hughes, M.~Lewis, and
  K.~Sycara, ``Individualized mutual adaptation in human-agent teams,''
  \emph{IEEE Transactions on Human-Machine Systems}, vol.~51, no.~6, pp.
  706--714, 2021.

\bibitem{hughes2020inferring}
D.~Hughes, A.~Agarwal, Y.~Guo, and K.~Sycara, ``Inferring non-stationary human
  preferences for human-agent teams,'' in \emph{2020 29th IEEE International
  Conference on Robot and Human Interactive Communication (RO-MAN)}.\hskip 1em
  plus 0.5em minus 0.4em\relax IEEE, 2020, pp. 1178--1185.

\bibitem{jaques2019social}
N.~Jaques, A.~Lazaridou, E.~Hughes, C.~Gulcehre, P.~Ortega, D.~Strouse, J.~Z.
  Leibo, and N.~De~Freitas, ``Social influence as intrinsic motivation for
  multi-agent deep reinforcement learning,'' in \emph{International Conference
  on Machine Learning}.\hskip 1em plus 0.5em minus 0.4em\relax PMLR, 2019, pp.
  3040--3049.

\bibitem{li2015communication}
S.~Li, W.~Sun, and T.~Miller, ``Communication in human-agent teams for tasks
  with joint action,'' in \emph{International Workshop on Coordination,
  Organizations, Institutions, and Norms in Agent Systems}.\hskip 1em plus
  0.5em minus 0.4em\relax Springer, 2015, pp. 224--241.

\bibitem{van2020learning}
E.~M. van Zoelen, A.~Cremers, F.~P. Dignum, J.~van Diggelen, and M.~M. Peeters,
  ``Learning to communicate proactively in human-agent teaming,'' in
  \emph{International Conference on Practical Applications of Agents and
  Multi-Agent Systems}.\hskip 1em plus 0.5em minus 0.4em\relax Springer, 2020,
  pp. 238--249.

\bibitem{williams1992simple}
R.~J. Williams, ``Simple statistical gradient-following algorithms for
  connectionist reinforcement learning,'' \emph{Machine learning}, vol.~8,
  no.~3, pp. 229--256, 1992.

\bibitem{fan2010modeling}
X.~Fan and J.~Yen, ``Modeling cognitive loads for evolving shared mental models
  in human--agent collaboration,'' \emph{IEEE Transactions on Systems, Man, and
  Cybernetics, Part B (Cybernetics)}, vol.~41, no.~2, pp. 354--367, 2010.

\bibitem{van2018effects}
W.~Van~Winsum, ``The effects of cognitive and visual workload on peripheral
  detection in the detection response task,'' \emph{Human factors}, vol.~60,
  no.~6, pp. 855--869, 2018.

\end{thebibliography}


 





\end{document}